\documentclass[letterpaper]{article}
\usepackage{aaai16}
\usepackage[FIGTOPCAP]{subfigure}
\usepackage{algorithm}
\usepackage{algorithmic}
\usepackage{amsmath}
\usepackage[pdftex]{graphicx}
\begin{document}
\title{Modeling Human Understanding of Complex Intentional Action \\ with a Bayesian Nonparametric Subgoal Model}

\author{
Ryo Nakahashi \\
%Computer Science and Artificial Intelligence Laboratory \\
Computer Science and Artificial Intelligence Laboratory \\
Massachusetts Institute of Technology, USA / \\
Sony Corporation, JAPAN\\
%Cambridge, MA  02139 \\
\texttt{Ryo.Nakahashi@jp.sony.com} \\
\And \hspace{.25in}
Chris L. Baker \and Joshua B. Tenenbaum \\
%Department of Brain and Cognitive Sciences \\
\hspace{.25in} Department of Brain and Cognitive Sciences \\
\hspace{.25in} Massachusetts Institute of Technology, USA \\
%Cambridge, MA  02139 \\
\hspace{.25in} \texttt{\{clbaker,jbt\}@mit.edu}
}
\maketitle
\begin{abstract}
\begin{quote}
Most human behaviors consist of multiple parts, steps, or subtasks. These
structures guide our action planning and execution, but when we observe others,
the latent structure of their actions is typically unobservable, and must be
inferred in order to learn new skills by demonstration, or to assist others in
completing their tasks. For example, an assistant who has learned the subgoal
structure of a colleague's task can more rapidly recognize and support their
actions as they unfold. Here we model how humans infer subgoals
from observations of complex action sequences using a nonparametric Bayesian
model, which assumes that observed actions are generated by approximately
rational planning over unknown subgoal sequences. We test this model with
a behavioral experiment in which humans observed different series of
goal-directed actions, and inferred both the number and composition of the
subgoal sequences associated with each goal. The Bayesian model predicts human
subgoal inferences with high accuracy, and significantly better than several
alternative models and straightforward heuristics. Motivated by this result, 
we simulate how learning and inference of subgoals can improve performance in an artificial user assistance task. The Bayesian model learns the correct subgoals from fewer observations, and better assists users by more rapidly and accurately inferring the goal of their actions than alternative approaches.
\end{quote}
\end{abstract}

\newcommand{\bvec}[1]{\mbox{\boldmath $#1$}}

\section{Introduction}
Human behavior is hierarchically structured. Even simple actions -- checking email, for example -- consist of many steps, which span levels of description and complexity: \emph{moving} fingers, arms, eyes; \emph{choosing} whether to use the mouse or keyboard; \emph{searching} for the email app among the others that are open, etc. Human behavior is also efficient: we attempt to perform each part of each action as swiftly and successfully as we can, at the least possible cost.

Classical models of how humans understand the actions of others (originating from the plan recognition literature~\cite{Schank1977,Kautz1986,Charniak1993,Bui2002}) seek to leverage this hierarchical structure by building in prior knowledge of others' high-level actions, tasks, and goals.
%More recent models perform statistical inference of the hierarchical structure of actions from data~\cite[e.g.]{Liao2004,Buchsbaum2015}. 
Behavioral evidence has shown that adults and even infants can learn simple hierarchical action structures from data, segmenting novel sequences along statistical action boundaries~\cite{Baldwin2008}. These abilities can be captured with a nonparametric action segmentation model~\cite{Buchsbaum2015}, which learns lists of actions that occur in sequence, and potentially cause observable effects.

However, purely statistical learning from data leaves implicit the intrinsic efficiency of intentional actions. For adults and infants, the assumption that others' actions are rational functions of their beliefs, desires, and goals is fundamental~\cite{Dennett1987,Gergely1995}.
Rather than simply memorizing repeated action sequences, people infer goals of complexity sufficient to rationalize these actions~\cite{Schachner2013}. A simple formalization of people's theory of rational, goal-directed action can be given in terms of approximately rational planning in Markov decision processes (MDPs), and human goal inferences can be accurately predicted using Bayesian inference over models of MDP planning (or inverse planning)~\cite{Baker2009}. This approach is closely related to recent plan recognition algorithms  developed in the AI literature~\cite[e.g.]{Ram2010}.

In this paper, we integrate hierarchically structured actions into the inverse planning framework using hierarchical MDPs~\cite{Sutton1999}. We consider scenarios in which agents pursue sequences of subgoals enroute to various destinations. Each destination can have multiple subgoal sequences, generated by a nonparametric Bayesian model. Together, a destination and subgoal sequence induce a hierarchical MDP, and agents' actions are assumed to be generated by approximately rational planning in this MDP. 

Representing agents' subgoal-based plans allows segmentation of behavior into extended chunks, separated by sparse subgoal boundaries, and can achieve greater generalization than purely statistical models by naturally capturing the context-sensitivity of rational action sequences, rather than having to learn new sequences for each context. Further, the model predicts deviations from rationality (with respect to a single subgoal) at subgoal boundaries -- a strong cue to hierarchical structure. This inductive bias should enable efficient learning of subgoal structure from small numbers of examples.

We present two experiments to test our model. The first is a behavioral experiment, in which human participants inferred the subgoal structure underlying series of action sequences. The second experiment is a simulation to show that the model is useful in an artificial user support task. The model first learns the subgoal structure of the task from a small number of observations. Then, the model infers a user's destination and subgoals from a partial action sequence, and attempts to assist the user to achieve a subset of the remaining subgoals. For each experiment, we compare the performance of our model with that of several natural alternatives.

\section{Computational Model}

\begin{figure}[ht]
	%	\centering
	\begin{tabular}{ll}
		\subfigure[Hierarchical MDP]{
			\includegraphics[clip,width=0.50\hsize]{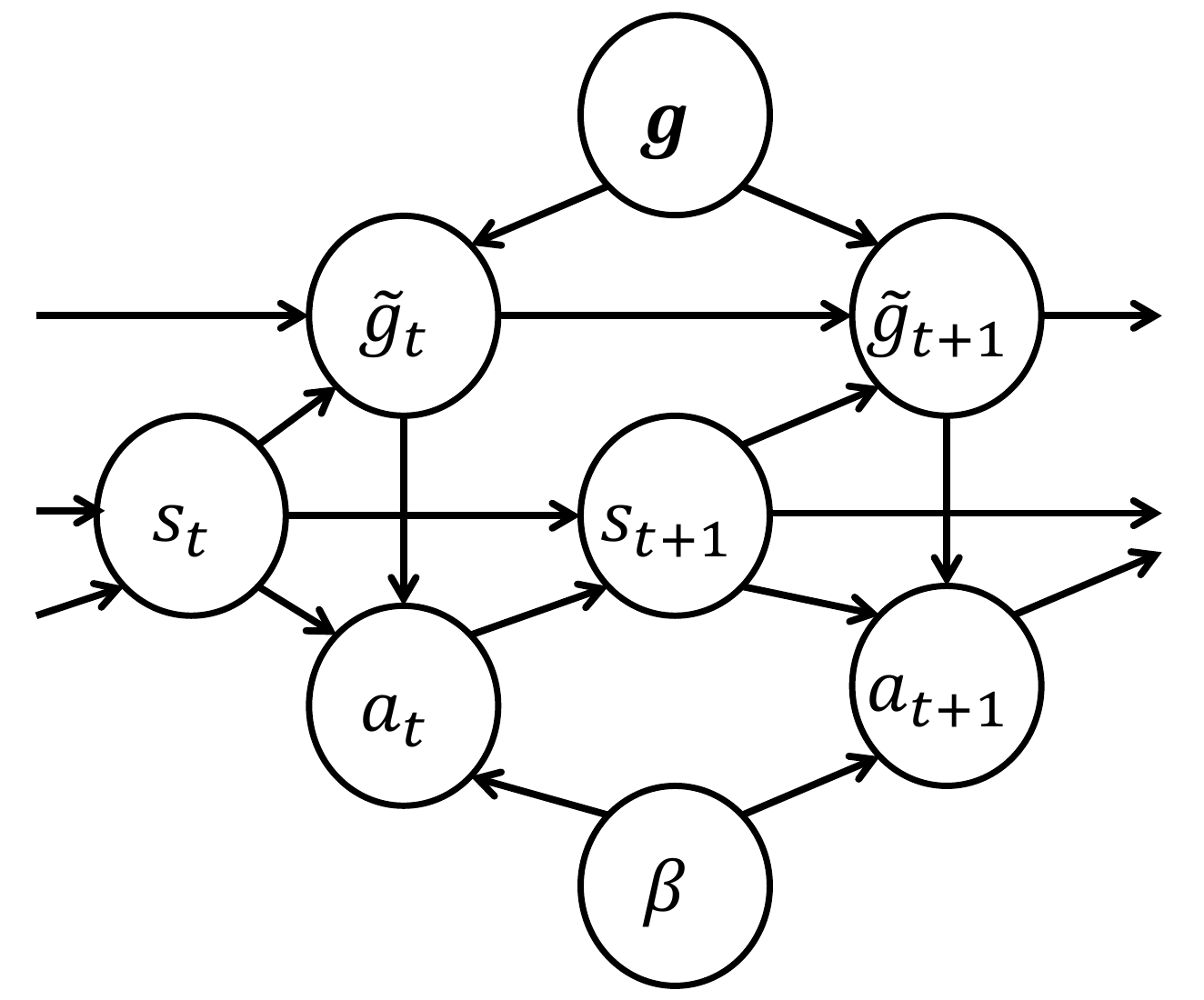}
			%\label{fig:graph-mdp}
		}
		\subfigure[Subgoal DP]{
			\includegraphics[clip,width=0.35\hsize]{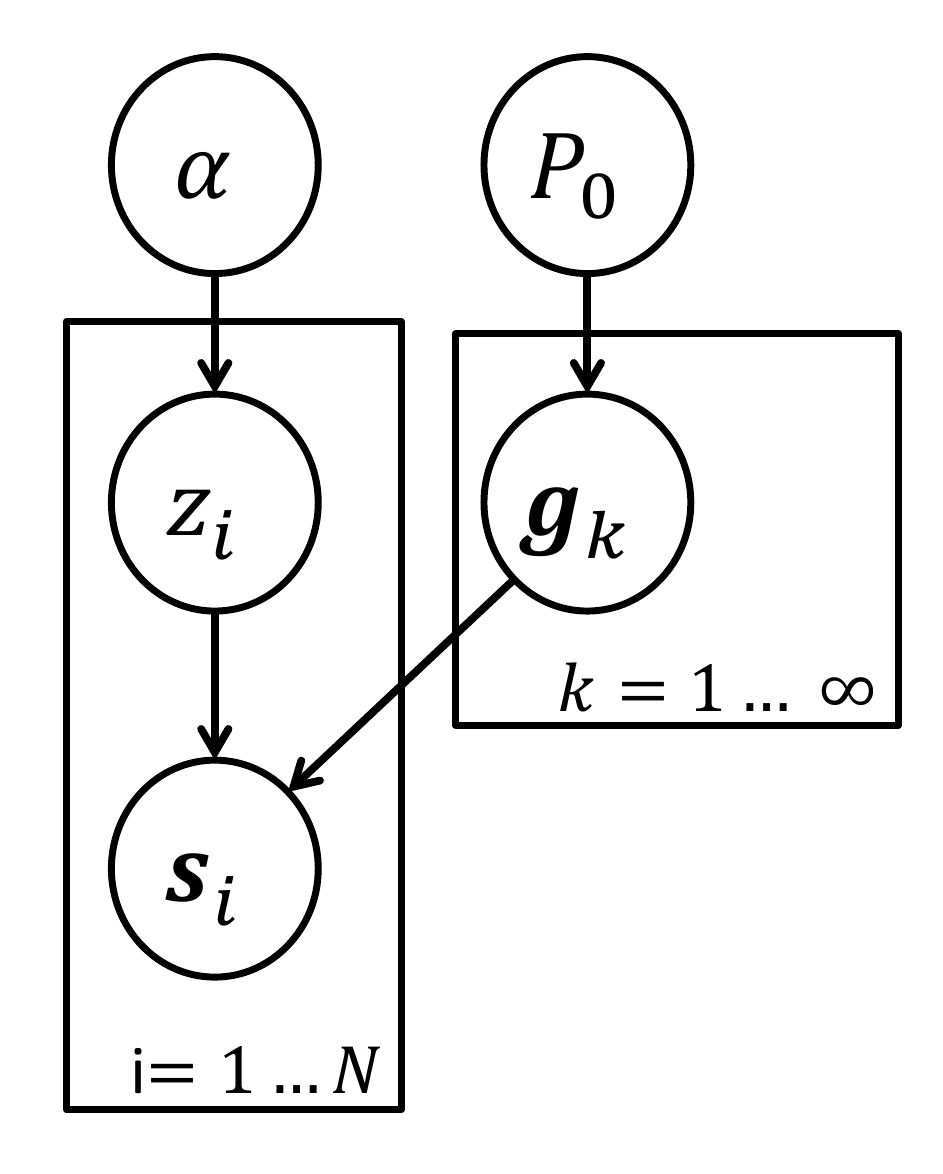}
			%\label{fig:graph-crp}
		}
	\end{tabular}
	\caption{Graphical models for our framework. (a) Hierarchical MDP planning model of state sequences. Agents select actions according to a probabilistic policy in the hierarchical MDP defined by the subgoal sequence \bvec{g}. $\beta$ is a parameter for soft-max action selection. (b) The Bayesian nonparametric subgoal model takes the form of a Dirichlet process (DP), in which each action sequence depends on a goal sequence sampled from a DP with concentration parameter $\alpha$.}
	\protect\label{fig:PP_str}
\end{figure}

Fig.~\ref{fig:PP_str} represents the structure of our model in terms of separate graphical models for the hierarchical MDP and  nonparametric subgoal models. These graphical models specify the structure of the joint distributions over actions, state sequences, and subgoal sequences. Our model represents the situational context in terms of a finite state space $\mathcal{S}$. The variable $\bvec{s}$ denotes a state sequence of length $T$, such that $s_t \in \mathcal{S}$ is the $t$th state in the sequence. The variable $\bvec{g}$ represents a sequence of $M$ subgoals. We denote the $m$th subgoal of $\bvec{g}$ as $g_m \in \mathcal{S}$. For convenience, we assume that $s_T = g_M = d$ is the destination. We denote the set of actions as $\mathcal{A}$, and the action executed in $s_t$ as $a_t \in \mathcal{A}$ (see Fig.~\ref{fig:PP_str}(a)).

The remainder of this section will first define the hierarchical MDP induced by a given destination and subgoal sequence, and derive the likelihood of a subgoal sequence, given an observed state sequence. Then we describe the nonparametric model of subgoal sequences, and a Markov chain Monte Carlo (MCMC) method for jointly inferring the number and composition of the subgoal sequences underlying a series of action sequences. Finally, we show how to use subgoal sequences learned from previous observations to predict the subgoals and destination of a novel partial action sequence.

\subsection{Subgoal sequence likelihood}
\label{sec:oneSeq}

Our hierarchical MDP formulation is closely related to the options framework for hierarchical reinforcement learning~\cite{Sutton1999}. For simplicity, we assume that actions and state transitions are deterministic, that each action incurs a cost of 2, that the discount factor $\gamma=1.0$, and that the destination yields a reward of 100 once all subgoals have been achieved.

The subgoal sequence $\bvec{g}$ is analogous to a set of options, with initiation and termination conditions that require each subgoal to be achieved in sequential order. In Fig.~\ref{fig:PP_str}(a), the variable $\tilde{g}_t$ keeps track of the current subgoal at time $t$. Assume the current subgoal is $\tilde{g}_t = g_m$; if the agent reaches the current subgoal, i.e., $s_t = \tilde{g}_t$, then $\tilde{g}_{t+1}\leftarrow g_{m+1}$ should be the next subgoal, otherwise the subgoal should stay the same ($\tilde{g}_{t+1}\leftarrow \tilde{g}_t$).

Based on this, an observed sequence $\bvec{s}$ can be divided into multiple segments corresponding to $\bvec{g}$. We define boundary $b_m$ to be the first timestep $t$ after $g_{m-1}$ is achieved: $b_m = \min(\{t|s_{t-1}=g_m\wedge t > b_{j-1} \})$; $b_0 = 1$. We write the boundary vector $\bvec{b}=\langle b_0,b_1,...\rangle$. If $\bvec{s}$ achieves all subgoals in $\bvec{g}$ in order, the length of $\bvec{b}$, $\dim(\bvec{b})$ should be $\dim(\bvec{g}) + 1$. Otherwise, $\bvec{s}$ does not satisfy $\bvec{g}$, so $P(\bvec{s} | \bvec{g})$ should be 0. The subgoal sequence likelihood is then:
\begin{eqnarray}
P(\bvec{s} | \bvec{g}) = 
\begin{cases}
\displaystyle \prod^{\dim(\bvec{g})}_{m=1} \prod^{b_m-1}_{t=b_{m-1}} P(s_{t+1}| s_t, g_m); \\ \quad (\text{if} \, \dim(\bvec{b})= \dim(\bvec{g}) + 1) \\
0 ; (\text{otherwise}), 
\end{cases}
\label{eq:ll_sb}
\end{eqnarray}
where $P(s_{t+1}|s_t,g_m) = \displaystyle\sum_{a_t \in A} P(s_{t+1}|s_t, a_t) P(a_t|s_t, g_m)$
is the marginal probability of the state transition from $s_t$ to $s_{t+1}$, integrating over actions. $P(a_t|s_t, g_m) \propto \exp(\beta Q_{g_m}(s_t, a_t))$ is a softmax policy of the local MDP state-action value function for subgoal $g_m$, based on the assumption that the observed agent plans approximately rationally, stochastically maximizing expected reward and minimizing cost.
 
A similar likelihood computation was used by~\cite{Michini2013} within an MCMC method for inferring subgoal sequences from user demonstrations. However, this approach focused on learning only one subgoal sequence from one action sequence; in the next section, we describe a nonparametric Bayesian model and MCMC methods for inferring multiple subgoal sequences, given a series of state sequences.

\subsection{Nonparametric subgoal inference}

We now consider inference of subgoal structure by observing multiple sequences for a certain destination. We denote the set of $N$ behavior sequences as $\bvec{s}_{1:N}$ and the $i$th sequence as $\bvec{s}_i$. We denote a set of $K$ subgoal sequences as $\bvec{g}_{1:K}$, and the $k$th sequence as $\bvec{g}_k$. The problem of nonparametric subgoal inference is to compute $P(\bvec{g}_{1:K} |\bvec{s}_{1:N})$ for an unbounded number of sequences~$K$.

\label{sec:gibbs}
We model the set of unknown subgoal sequences using a nonparametric Bayesian model, which allows us to consider an unbounded number of subgoal sequences for each destination. We use the Dirichlet process (DP) to express the distribution over subgoal sequences, following~\cite{Buchsbaum2015}. A graphical model of our DP model is shown in Fig.~\ref{fig:PP_str}(b). We use the Chinese Restaurant Process (CRP) representation to efficiently draw samples from the DP. First, the CRP selects a ``table'' for each observation $\bvec{s}_i$, conditioned on all previous table assignments and the concentration parameter $\alpha_0$. $z_i$ is the index of the table assigned to state sequence $\bvec{s}_i$. Next, for each CRP table, a subgoal sequence is sampled  from the base distribution $P_0$, and $\bvec{g}_k$ denotes the subgoal sequence associated with the $k$th table. The state sequence $\bvec{s}_i$ is then generated given its associated subgoal sequence.% We choose a base distribution $P_0$ whose probability for one subgoal sequence can be computed analytically. 

We use a MCMC method to compute the posterior probability over subgoal sequences, specifically, Gibbs sampling.
Gibbs sampling allows us to approximate the complex DP distribution by inducing a Markov chain over samples from the CRP. Gibbs sampling over the CRP is a standard MCMC algorithm for DP inference~\cite{Neal2000}.
Algorithm~\ref{alg:gibbs} is an overview of our algorithm. As an initialization step, we assign a different table for each state sequence, and draw subgoal sequences from the conditional distribution over subgoal sequences given the state sequence assigned to each table. We then repeat the table re-assignment step (resampling the table for each state sequence) and the parameter re-assignment step (drawing the subgoal sequences from the conditional distribution over subgoal sequences, given all sequences assigned to a table). For the table re-assignment step, we calculate the probability $P(z_i = k| z_{-i}, \bvec{s}_i)$ to assign sequence $\bvec{s}_i$ to table $k$  according to standard Gibbs sampling for the CRP:
% \begin{eqnarray}
\begin{multline}
P(z_i = k| z_{-i}, \bvec{s}_i) =\\
\begin{cases}
\displaystyle \frac{n_{-i,k}}{N-1+\alpha} P(\bvec{s}_i | \bvec{g}_k) \\ ({\rm If\ } k = z_j\ {\rm for\ some\ } i \neq j) \\
\displaystyle \frac{\alpha}{N-1+\alpha} \int P(\bvec{s}_i|\bvec{g}) P_0(\bvec{g}) d\bvec{g} \\ ({\rm If\ } k\neq z_j\ {\rm for\ all\ } i\neq j),
\end{cases} 
\label{eq:gibbs}
\end{multline}
% \end{eqnarray}
where $z_{-i}$ denotes table assignments, excluding sequence $i$, and $n_{-i,k}$ denotes the number of sequences assigned to table $k$, excluding sequence $i$. 

However, for our problem, $P(\bvec{s}_i|\bvec{g})$ is the MDP likelihood of a subgoal sequence. Because this is a non-conjugate distribution, we cannot integrate this equation analytically. If the environment is small, we can enumerate all of \bvec{g}, and compute it directly as in the previous section. In large environments we must use an approximate method to choose a new table; some techniques are described by~\cite{Neal2000}. In the parameter re-assignment step, we draw the subgoal sequence from the posterior over subgoal sequences, given all sequences assigned to a table. Assume $\bvec{s}_{1:l}$ are the sequences assigned to table $k$. The distribution to draw a new subgoal sequence $\bvec{g}$ for table $k$ should be $P(\bvec{g} | \bvec{s}_{1:l})$. This probability for each subgoal sequence can be calculated as follows:
\begin{eqnarray}
\label{eq:ll_sb_multi}
P(\bvec{g} | \bvec{s}_{1:l}) \propto P_0(\bvec{g}) P(\bvec{s}_{1:l} |
\bvec{g})  = P_0(\bvec{g}) \prod^{l}_{i=1} P(\bvec{s}_i | \bvec{g})
%\nonumber P(\bvec{g} | \bvec{s}_{1:l}) \propto P_0(\bvec{g}) \prod^{l}_{i=1} P(\bvec{g} | \bvec{s}_i) 
%& \propto & P_0(\bvec{g}) \prod^{l}_{i=1} \frac{P(\bvec{s}_i | \bvec{g})}{P_0(\bvec{g})} \\ & = & \frac{\prod^{l}_{i=1}P(\bvec{s}_i | \bvec{g})}{(P_0(\bvec{g}))^{l-1}}
\end{eqnarray}

At the end of each step of the loop, we count the number of subgoal sequences for each state sequence. We represent the number of times that $\bvec{g}$ is assigned any state sequence as ${\bf c}(\bvec{g})$. The normalized count corresponds to $P(\bvec{g} \in \bvec{g}_{1:K} |\bvec{s}_{1:N})$.

\begin{algorithm}
	\caption{Subgoal inference}         
	\label{alg:gibbs}                          
	\begin{algorithmic}
		\FOR {$i=1$ to $N$} 
		\STATE $z_i = i; \bvec{g}_i \sim P(\bvec{g}_i | \bvec{s}_i)$ // Initialize Step (See, Eq.  \ref{eq:ll_sb})
		\ENDFOR
		\FOR {$r=1$ to $repeat$}
		\FOR {$i=1$ to $N$}
		\STATE $z_i \sim P(z_i| z_{-i}, \bvec{s}_i)$ // Table Re-assign Step (See, Eq. \ref{eq:gibbs})
		\IF {$z_i$ is index for new table}
		\STATE	$\bvec{g}_{z_i} \sim P(\bvec{g}_{z_i} | \bvec{s}_i)$ // Parameter Initialize for new table (See, Eq. \ref{eq:ll_sb})
		\ENDIF
		\ENDFOR
		\FOR {$k\in \{z_1, z_2,...\}$}
		\STATE $\displaystyle \bvec{g}_k \sim P(\bvec{g}_k|\{\bvec{s}_i | 1 \leq i \leq N, z_i = k\}) $  // Parameter Re-assign Step (See, \ref{eq:ll_sb_multi})
		\ENDFOR
		\FOR {$k\in \{z_1, z_2,...\}$}
		\STATE ${\bf c}(\bvec{g}_{k}) \leftarrow {\bf c}(\bvec{g}_{k}) + 1$
		\ENDFOR
		\ENDFOR
		\FORALL {$\{\bvec{g} | {\bf c}(\bvec{g}) > 0\}$}
		\STATE $P(\bvec{g} \in \bvec{g}_{1:K} | \bvec{s}_{1:N}) \leftarrow {\bf c}(\bvec{g}) / (repeat)$
		\ENDFOR
		\STATE {\bf output} $P(\bvec{g} \in \bvec{g}_{1:K} |\bvec{s}_{1:N})$
	\end{algorithmic}
\end{algorithm}

\section{Experiments}

\begin{figure}[ht]
	%	\centering
	\begin{tabular}{ccc}
		\subfigure[Warehouse \newline environment]{
			\includegraphics[clip,width= 0.28\hsize]{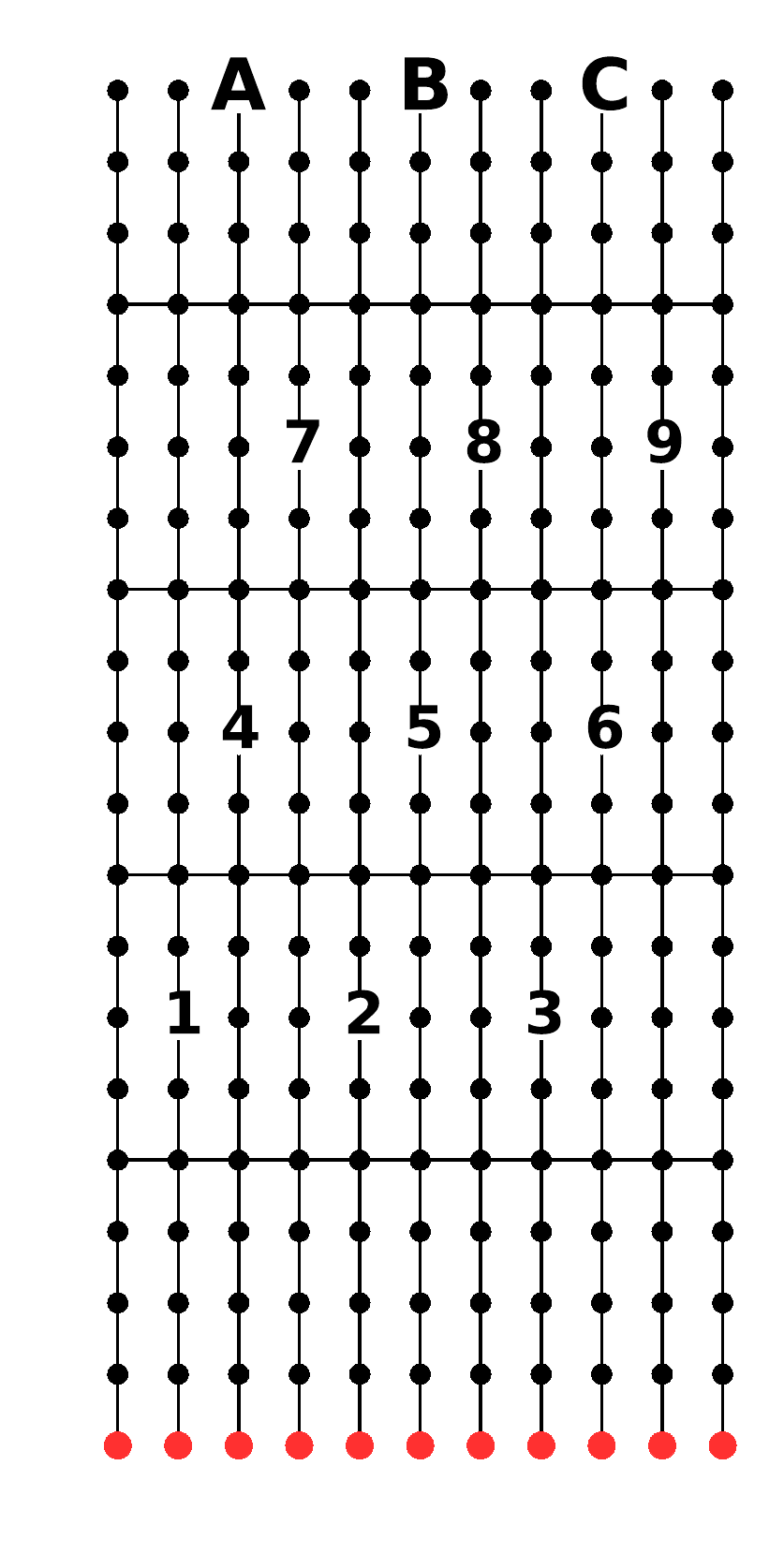}
			%\label{fig:exp_env_none}
		}
		\subfigure[Example \newline sequence]{
			\includegraphics[clip,width= 0.28\hsize]{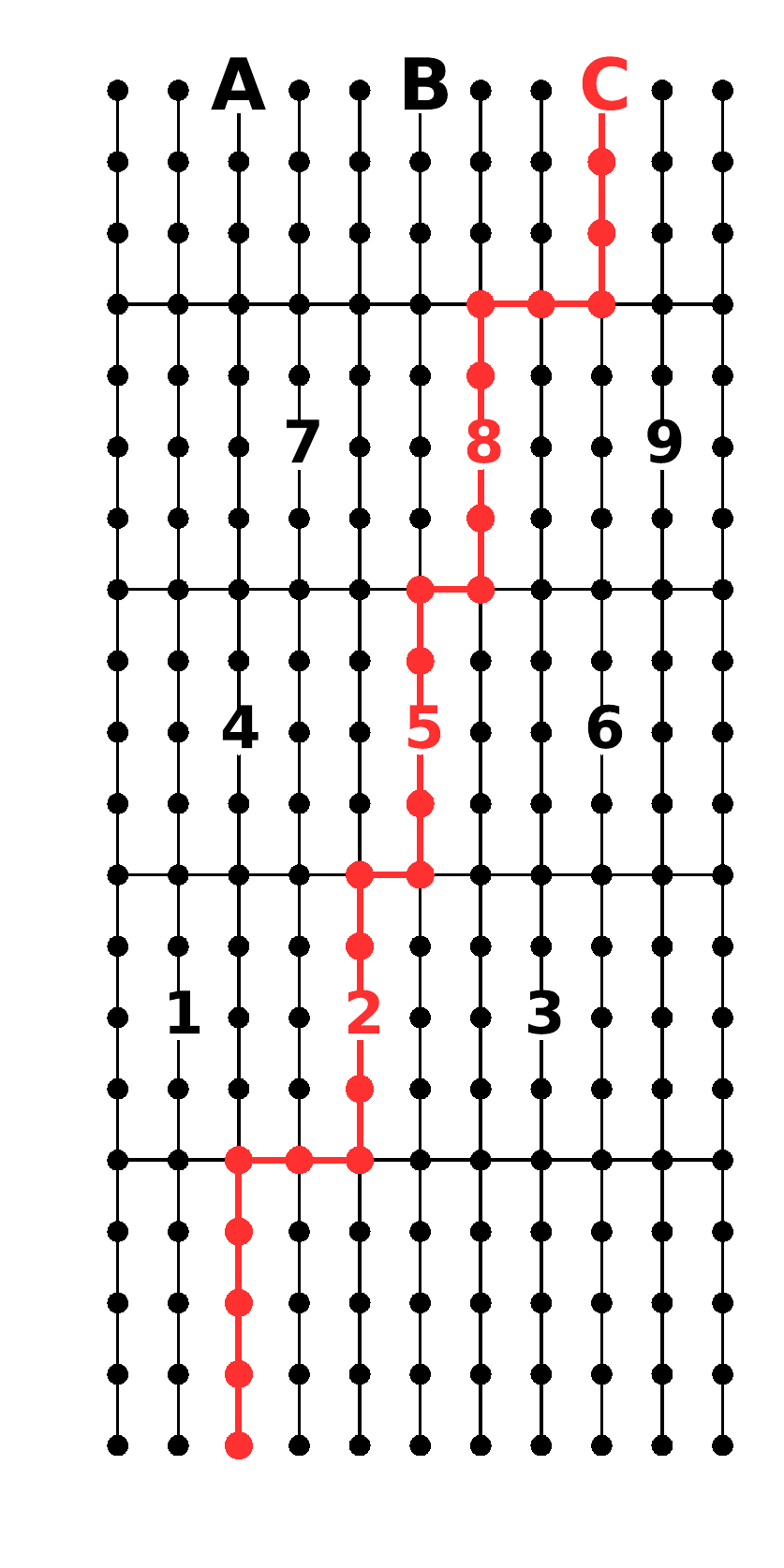}
			%\label{fig:exp_env_path}
		}
		\subfigure[Worker-Helper \newline environment]{
			\includegraphics[clip,width= 0.28\hsize]{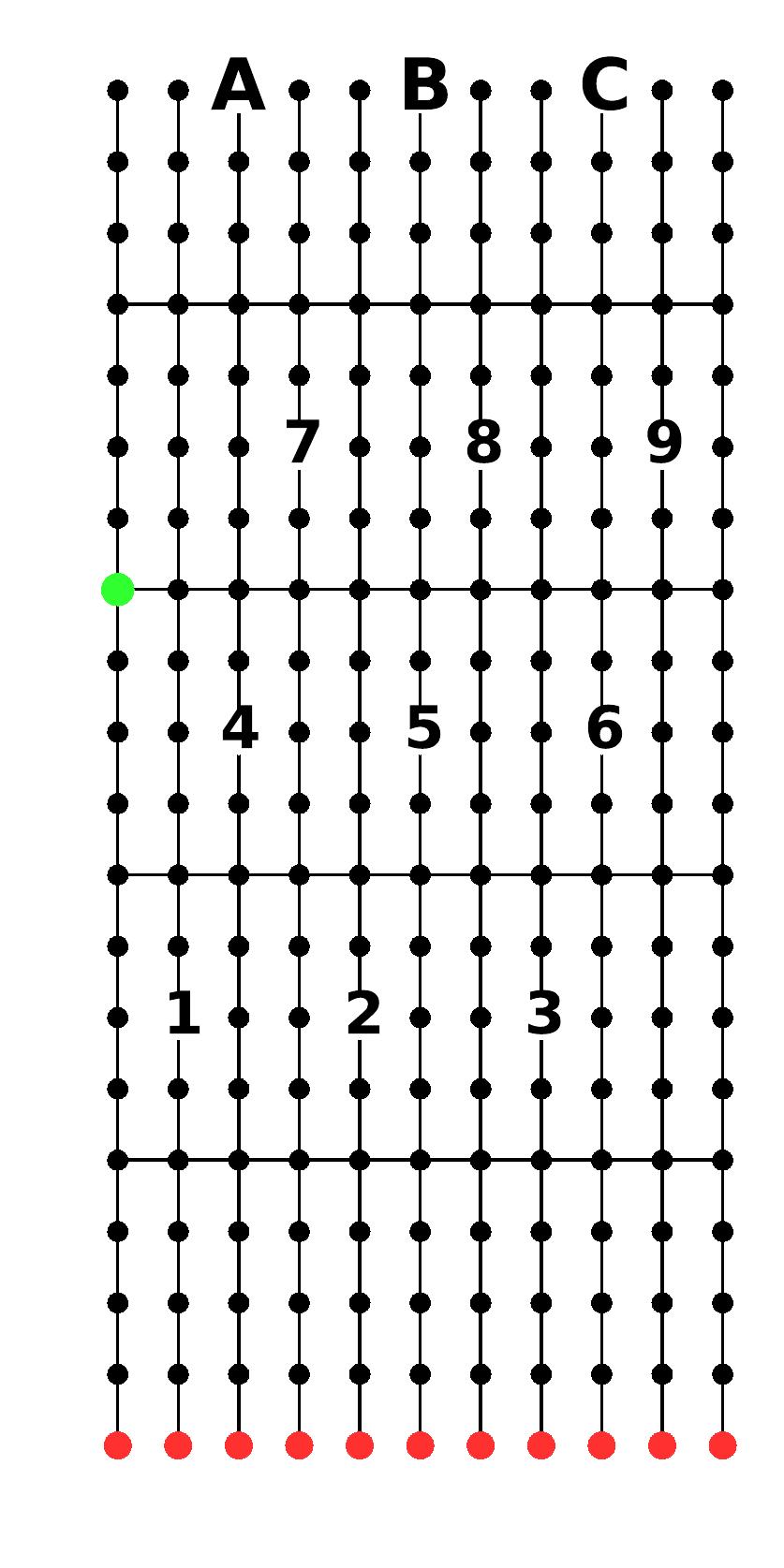}
			%\label{fig:exp_env_path}
		}
	\end{tabular}
	\caption{(a) Experimental scenario. A, B, and C mark possible destinations, and numbers 1-9 mark items that workers must deliver. Red points mark potential starting locations.(b) Example agent behavior. (c) Environment of Experiment 2. Green point is the starting location of the Helper.}
	\label{fig:warehouse}
\end{figure}

Our two experiments presented a ``warehouse'' scenario involving the delivery of various items to destinations in the environment shown in Fig.~\ref{fig:warehouse}. The warehouse has three delivery destinations: A, B, and C. There are nine potential items to be delivered, marked by numbers 1-9. The items are arranged into three rows, and for each delivery, one item can be delivered from each row.

Each job in the warehouse has a specific destination, and several possible ``item lists'' to deliver. An item list consists of either one, two, three items that must be delivered to the destination. For each delivery, one of these item lists is selected by the warehouse scheduler. In addition to items from the current item list, workers are encouraged to pick up ``Add-on'' items , but only if this won't increase the number of steps on their path to the destination. For example, if the item list for Fig.~\ref{fig:warehouse}(b) is [2,8], item 5 is a good Add-on item for that delivery, because item 5 is the only additional item that does not require additional steps to obtain, given the start point, item list, and destination.  There is a direct correspondence between delivery destinations and item lists in this setting and destinations and subgoals as represented by our model.

\subsection{Experiment 1: Subgoal inference}

\subsubsection{Participants}
We recruited participants for this study using Amazon Mechanical Turk. Participants were 29 adults located in the USA (Male 13, Female 14, Unknown 2). Mean age was 32 years old.

\subsubsection{Procedure}
First, in a training phase, subjects learned to perform an example warehouse job to learn the rules of the scenario.
In each trial of the testing phase, participants saw 8 paths for a particular job, and inferred how many item lists the job had, and which items were on each list.

\subsubsection{Stimuli}
We prepared 22 jobs for the testing phase, including a range of subgoal structures of varying complexity and difficulty. First were jobs with one item list, involving 1, 2, or 3 subgoals. Second were jobs with two item lists,  involving 1, 2, or 3 subgoals for both item lists, or 1 and 3 subgoals for each item list, respectively. Some jobs presented two types of action sequences. In one type, the action sequence could reach all subgoals without deviating from the shortest path to the destination. These cases were difficult in that they did not provide direct evidence for any subgoal. In the other type, the action had to make a detour to reach the subgoal, which provided stronger evidence for that subgoal.

\subsubsection{Modeling}
In our MDP representation of the environment, $\mathcal{S}$ corresponds to the agent location, and $\mathcal{A}$ is the set of movements (up, left, and right). We used a uniform distribution for the prior over subgoal sequences $P_0(\bvec{g})$ and set $\beta = 6$. For the nonparametric Bayesian model, we set the concentration parameter $\alpha_0=0.015$. We used 5,000 iterations of Gibbs sampling, with 1,000 steps burn-in time to reach the stationary distribution.

\paragraph{Alternative models}
We prepared 3 alternative models to compare with our Bayesian nonparametric model: {\it Independent model}, {\it Logical possibility model}, and {\it Copy model}. To explain each alternative model, we consider the case in which the model observes $N$ sequences $\bvec{s}_{1:N}$.

The Independent model is most similar to our approach. The only difference is that this model does not use the CRP, but instead calculates the posterior for each sequence $P(\bvec{g}_i | \bvec{s}_i)$ independently. The joint probability for all sequences is given by: $P(\bvec{g}_{1:N} | \bvec{s}_{1:N}) = \prod_{i=1}^N{P(\bvec{g}_i | \bvec{s}_i)}$. The marginal probability for each subgoal sequence is given by: $P(\bvec{g} \in \bvec{g}_{1:N} | \bvec{s}_{1:N}) = \sum_{\{\bvec{g}_{1:N}|\bvec{g} \in \bvec{g}_{1:N}\}}P(\bvec{g}_{1:N} | \bvec{s}_{1:N})$.

The Logical possibility model is a more heuristic approach, which computes the proportion of observations in which a subgoal sequence is present within the action sequence. Formally, $P(\bvec{g} \in \bvec{g}_{1:K} | \bvec{s}_{1:N}) = |\{i| 1 \leq i \leq N \land \bvec{g} \subset \bvec{s}_i \}| / N$.

The Copy model is the simplest alternative, but perhaps the most intuitive. It assigns probability 1 to the subgoals which have the highest likelihood for each observed sequence. In other words, the inferred subgoal sequence is the maximal subgoal sequence that is included by an observed action sequence. If we represent such a subgoal sequence $MaxSubgoal(\bvec{s})$, the copy model can be written: $P(\bvec{g} \in \bvec{g}_{1:K} | \bvec{s}_{1:N}) = 1$ if $\bvec{g} = MaxSubgoal(\bvec{s}_i)$ for some $i$, otherwise 0.

\subsubsection{Results}
To compare human behavioral results with those of our models, we compute the proportion of participants who selected each item list for each trial. Model predictions were based on the marginal probability that each item list generated at least one path.

Table~\ref{tbl:exp1corr} compares the correlation of the Bayesian nonparametric subgoal model and our three alternative models with human inferences. The Bayesian nonparametric model correlates very strongly with human inferences. The other models also correlate positively with human inferences, but all are substantially worse than the Bayesian nonparametric model.
\begin{table}[ht]
{\small
  \begin{tabular}{|c||c|c|c|c|}\hline 
    Model & CRP & Independent & LP & Copy \\ \hline
    Correlation & 0.973 & 0.644 & 0.267 & 0.495 \\ \hline
    % \multirow{2}{*}{Model} & \multirow{2}{*}{CRP} & \multirow{2}{*}{Independent} & Logical & \multirow{2}{*}{Copy}  \\
    % & & & Possibility &  \\ \hline
    % Pearson & \multirow{2}{*}{0.973} & \multirow{2}{*}{0.644} & \multirow{2}{*}{0.267} & \multirow{2}{*}{0.495} \\ 
    % Correlation & & & & \\\hline
  \end{tabular}
}
  \caption{Pearson correlation between human inferences and computational models. LP stands for ``Logical Possibility''.}
    \label{tbl:exp1corr}
\end{table}

\begin{figure}[htbp]
	\includegraphics[clip,width= 0.9\hsize]{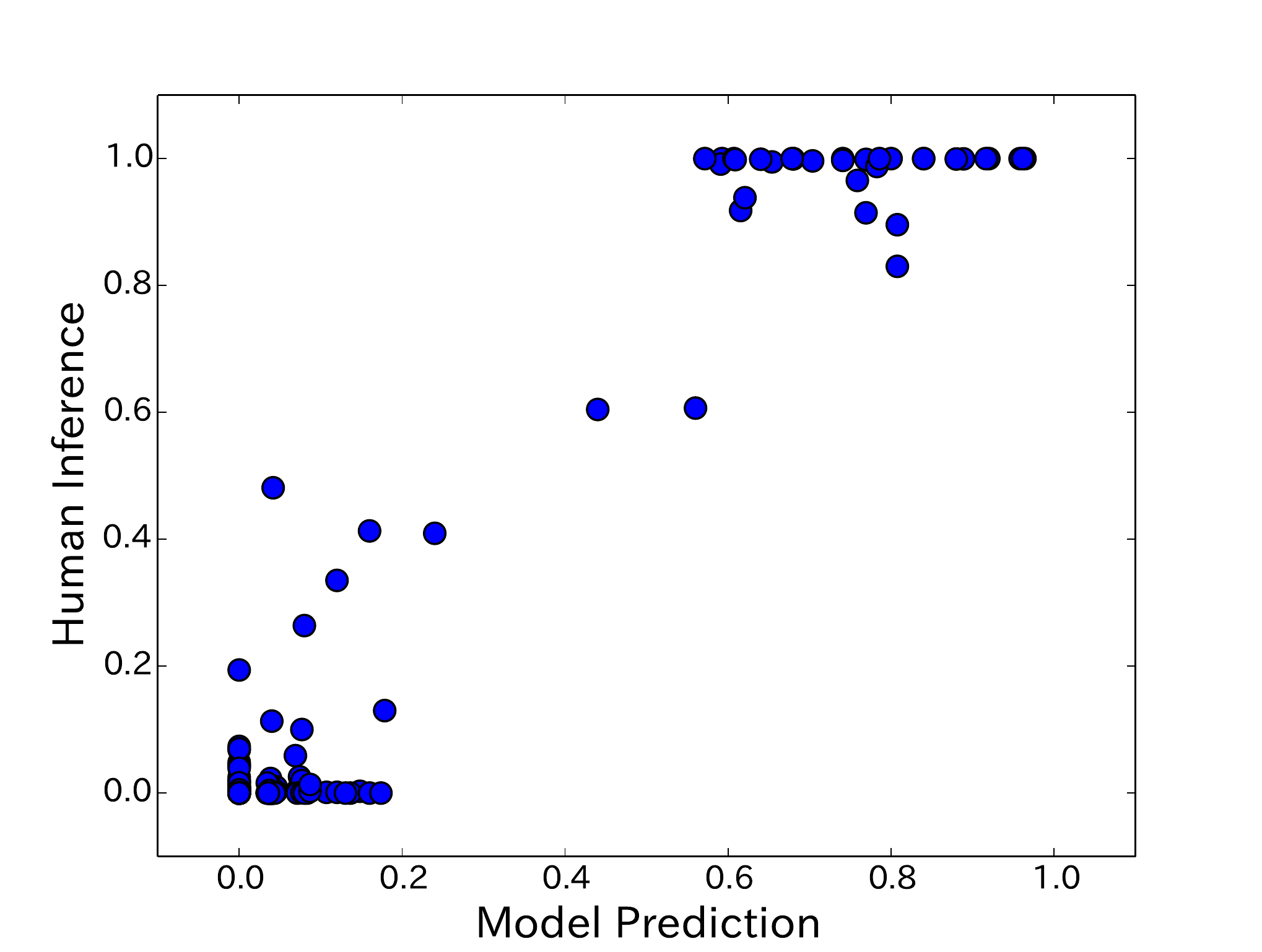}
	\caption{Scatter plot of human inferences versus Bayesian subgoal model predictions (r = 0.973).}
	\label{fig:exp1scatter}
\end{figure}

Fig.~\ref{fig:exp1scatter} plots human inferences against the Bayesian nonparametric model predictions for all trials and subgoal sequences. This shows that our model can almost perfectly predict whether humans will infer that a certain item list is included in each job. For a small number of trials, humans are somewhat uncertain whether to include a particular item list, and the model captures this uncertainty as well.

\begin{figure}[htb]
    \begin{tabular}{c}
        \subfigure[Results for job 2/22]{
        	\includegraphics[width=0.91\hsize]{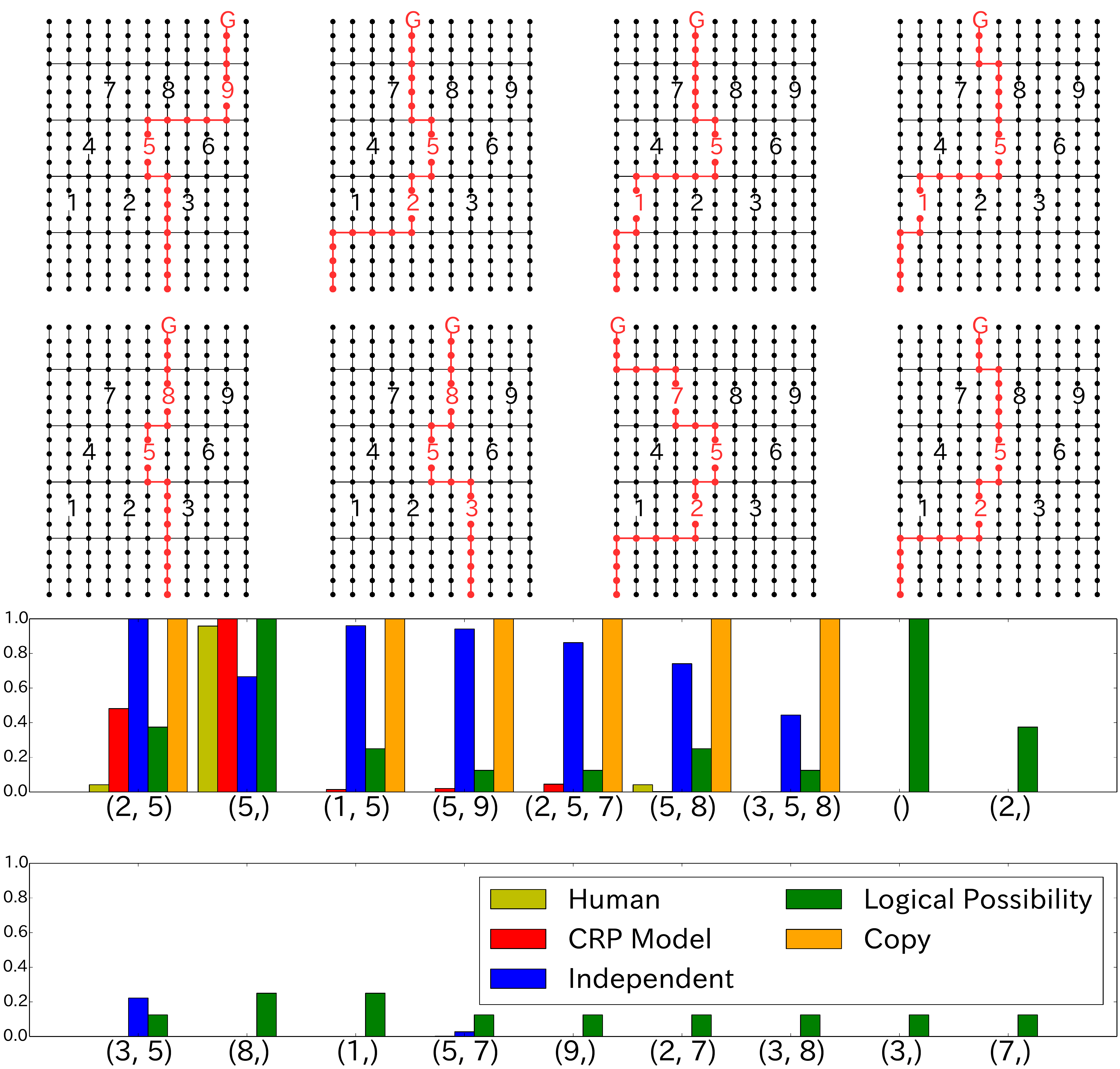}
        	%\label{fig:exp1_agent_2}
    	}
    \end{tabular}
    \begin{tabular}{c}
        \subfigure[Results for job 9/22]{
    	    \includegraphics[width=0.91\hsize]{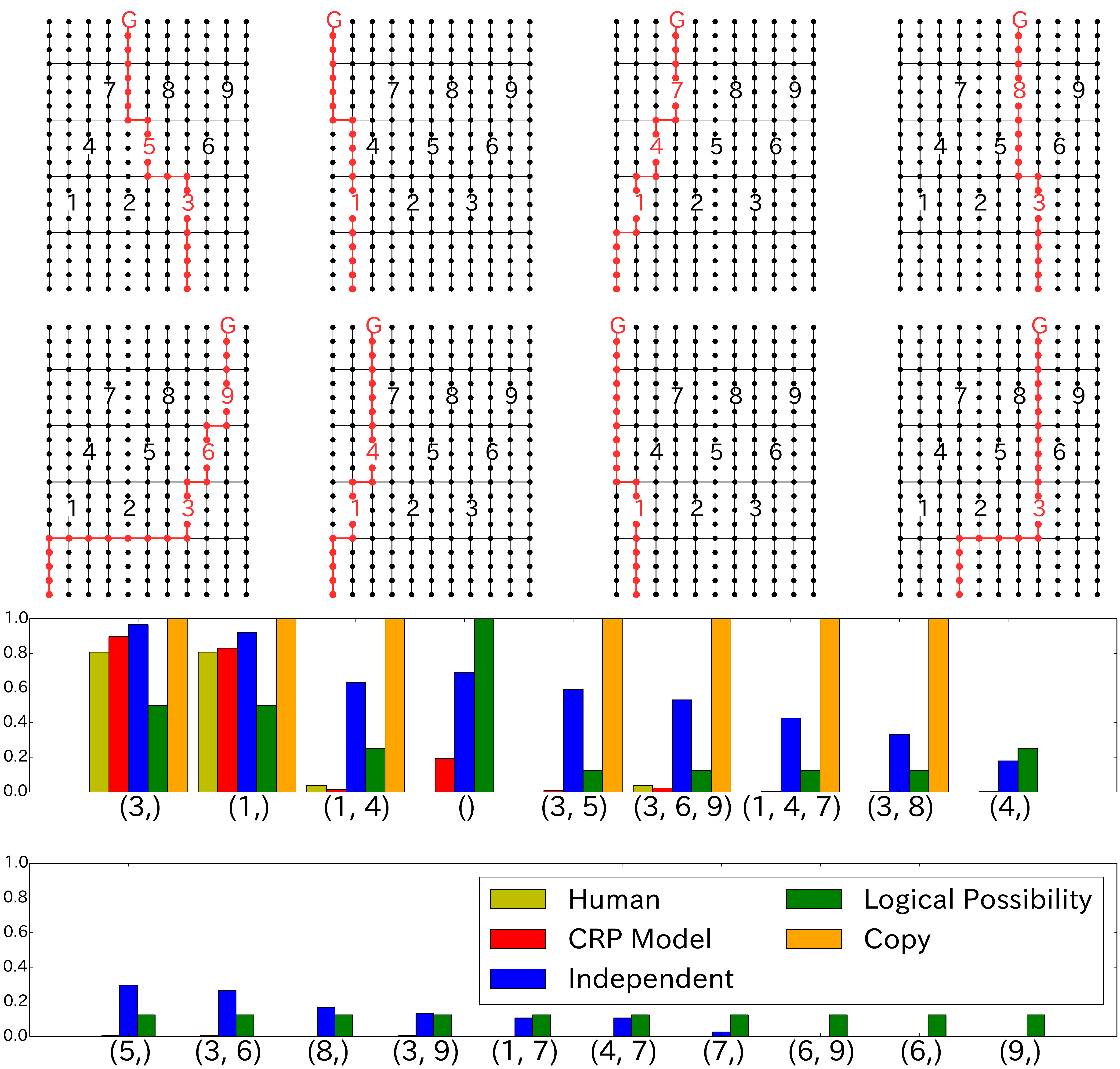}
    	    %\label{fig:exp1_agent_9}
    	}    
    \end{tabular}
    \caption{Example results for two jobs. (a) Job with item list [5]; humans and CRP model infer [5] is the most probable subgoal sequence. Alternative models fail to predict human judgments. (b) Job with two item lists: [1] and [3]. Humans and CRP model correctly infer these two subgoal sequences. Once again, the alternative models fail to predict human judgments.}
    \label{fig:exp1_agents}
\end{figure}

Finally, we compare our model predictions with human inferences for two specific jobs. In Fig.~\ref{fig:exp1_agents}(a), each path includes item 5, but there are no paths which include only this item. Humans and the Bayesian nonparametric model infer that [5] is the most likely subgoal sequence, based on the systematic deviation from the shortest path to the destination to reach this item. All other models fail in revealing ways, inferring more item lists than are necessary to explain the data.
In Fig.~\ref{fig:exp1_agents}(b), there are two item lists. Although every path is an efficient route to the destination, without inferring an item list, humans and our model can infer these two subgoal sequences correctly by assessing the probability that these items are selected intentionally rather than by coincidence. Our alternative models once again fail to capture people's judgments.

\subsection{Experiment 2: User support using subgoal inference}

\subsubsection{Task definition}
We call this the {\bf Worker-Helper Task}. The main environment is the same as in Experiment 1, but there are two types of agents. One type is {\bf Worker}, whose actions are structured identically to those presented in Experiment 1. The Worker's job remains the same: to deliver an item list to a destination. The other type is {\bf Helper}, which must support Workers in achieving their goals. The task of the Helper is to learn the structure of the Worker's jobs, then use this to help the Worker complete a job in progress by retrieving an item from the item list of the Worker, and delivering it to the destination. Helpers begin each trial on the left side of the warehouse, midway between the second row of items (4, 5 and 6) and the third row of items (7, 8 and 9). Fig.~\ref{fig:warehouse}(c) illustrates this task, with the start point of the Helper marked by a green dot.

\subsubsection{Collaboration protocol}
On each trial, the Worker randomly chooses a destination (A, B, or C) and an item list associated with that destination. The set of item lists for each destination is fixed and known to the Worker, but not the Helper. Next, the Worker plans a path to achieve their goals and begins to move through the warehouse. The Helper observes the Worker's behavior and decides which target item to retrieve by inferring the Worker's subgoal sequence (due to the Helper's starting location, the target item will always be in the third row, i.e., item 7, 8, or 9). Once the Helper decides their target item, the Helper shows this target to the Worker and begins to move. After the Worker observes the Helper's target item, the Worker re-plans its path under the assumption that the Helper will get the target item. After the Helper gets the target item, the Helper moves toward the destination inferred by observing the Worker's path. When the Helper cannot decide on a target item, they do nothing.

\subsubsection{Modeling}
\newcommand{\argmax}{\mathop{\rm arg~max}\limits}
We assume that workers take the optimal actions, given their subgoal sequence, i.e., $a_t = \argmax_{a_t} Q^{\pi^*}_{\bvec{g}}(s_t, a_t)$. The Helper estimates the marginal probability that each item is included in the subgoal sequence which the Worker is currently following. The Helper computes this probability based on the partial path of the Worker $s_{1:t}$, and $n$ previously observed paths $\bvec{s}_{1:n}$, under the assumption that the Worker plans approximately rationally, with softmax parameter $\beta=2$. When the probability of a target exceeds a certain value, the Helper decides on this item. The Worker then re-plans by removing the target item and the Helper begins to take optimal actions, as does the Worker. 

\paragraph{Probability of target item and destination}
The marginal probability of target item $g'$, given $s_{1:t}$ and $\bvec{s}_{1:n}$ is: $P(g' | s_{1:t},\bvec{s}_{1:n}) = \sum_{d} \sum_{\{\bvec{g}|g' \subseteq \bvec{g}\}} P(s_{1:t}|\bvec{g},d) P(\bvec{g} \in \bvec{g}_{1:K} |\bvec{s}^d_{1:n})$, where $\bvec{s}^d_{1:n}$ is the subset of previously observed paths with destination $d$. $P(\bvec{g} \in \bvec{g}_{1:K} |\bvec{s}^d_{1:n})$ is calculated using the subgoal inference method in Section 2. $P(s_{1:t} | \bvec{g}, d)$ can be computed using Eq. \ref{eq:ll_sb}, but without the constraint that $\dim( \bvec{g}) = \dim(\bvec{g}) + 1$.
The marginal probability of destination $d$, given $s_{1:t}$ and $\bvec{s}_{1:n}$ is: $P(d | s_{1:t},\bvec{s}_{1:n}) \propto \sum_{\bvec{g}} P(s_{1:t}|\bvec{g},d) P(\bvec{g} \in \bvec{g}_{1:K} |\bvec{s}^d_{1:n})$.

\paragraph{Alternative models}
We use the same alternative models as in Experiment 1, and a common collaboration protocol (described above) for each subgoal inference model. We also add two new models. One is No Helper, which means the Helper has no subgoal knowledge. As a result, the Helper does nothing. The other is the Ground Truth model, which means the Helper knows the correct subgoal. It is a benchmark to measure the benefit of the Bayesian model.

\subsubsection{Test method}
We tested three natural types of subgoal settings. In the first, each of the three destinations has one subgoal sequence, which consists of either 7, 8, or 9. In the second, each of the three destinations has one subgoal sequence, which consists of either item 4, 5, or 6 and either item 7, 8, or 9. In the third, each of the three destinations has two subgoal sequences which consist of either item 7, 8, or 9. We generate every possible combination of items for each subgoal setting, then we learn many subgoal structures and evaluate the performance for each subgoal structure.

\paragraph{Evaluation for one subgoal structure}
We generate a number of sequences of Workers from random start points for each destination. We then compute subgoal inferences for each model using the sequences. We then execute the collaborative task 99 times for each subgoal inference (11 points times 9 trials), and evaluate the performance of Workers in achieving their destination. We repeat these tests 5 times for each subgoal structure. We use the following score to measure performance. If the Worker achieves their goal, they score 100 points, but each action costs 2 points. This directly corresponds to the MDP reward and cost settings in Experiment 1.

\subsubsection{Results}
\begin{figure}[ht]
(a)
\includegraphics[clip,width= 0.9\hsize]{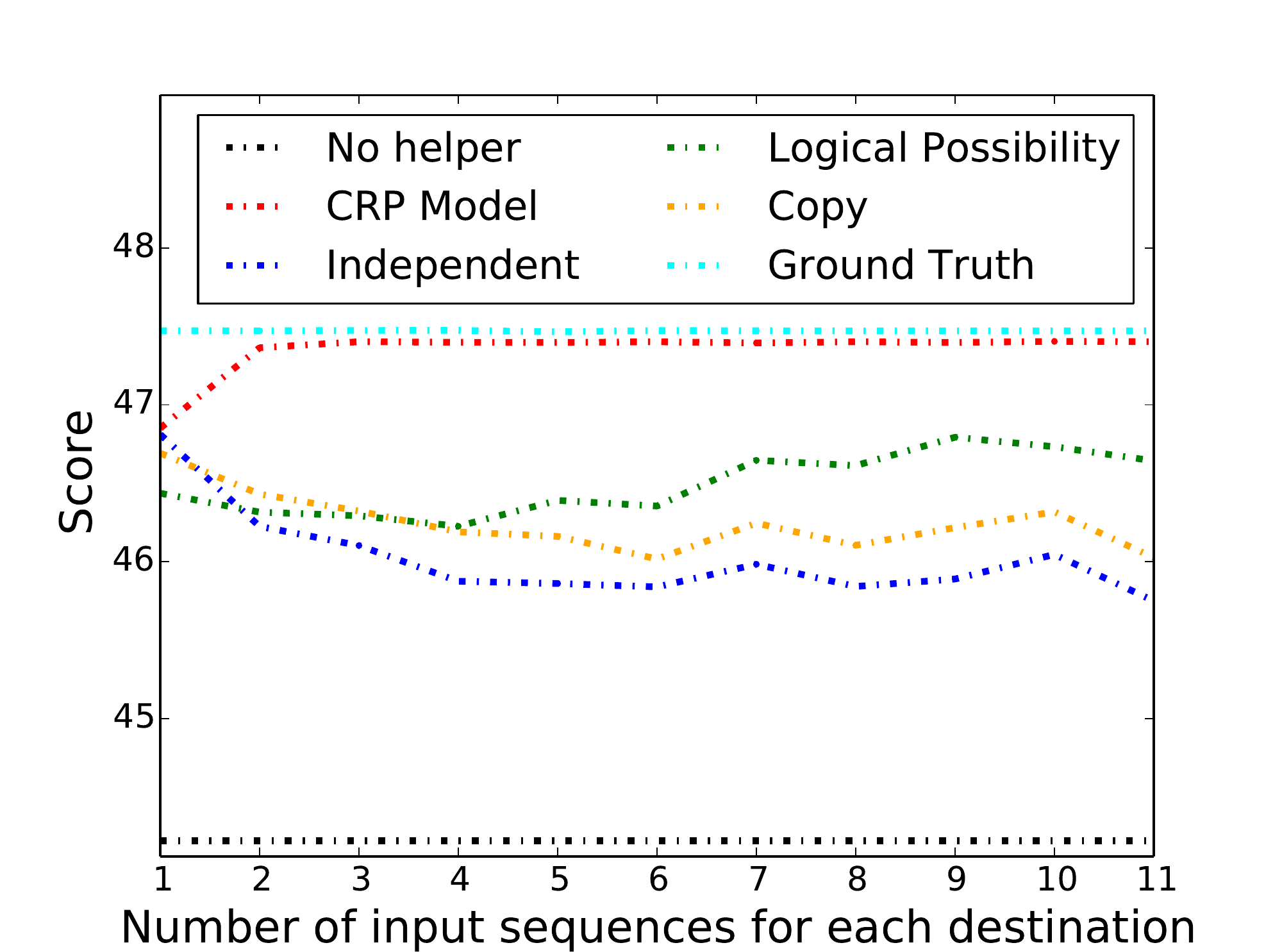}\\
(b)
\includegraphics[clip,width= 0.9\hsize]{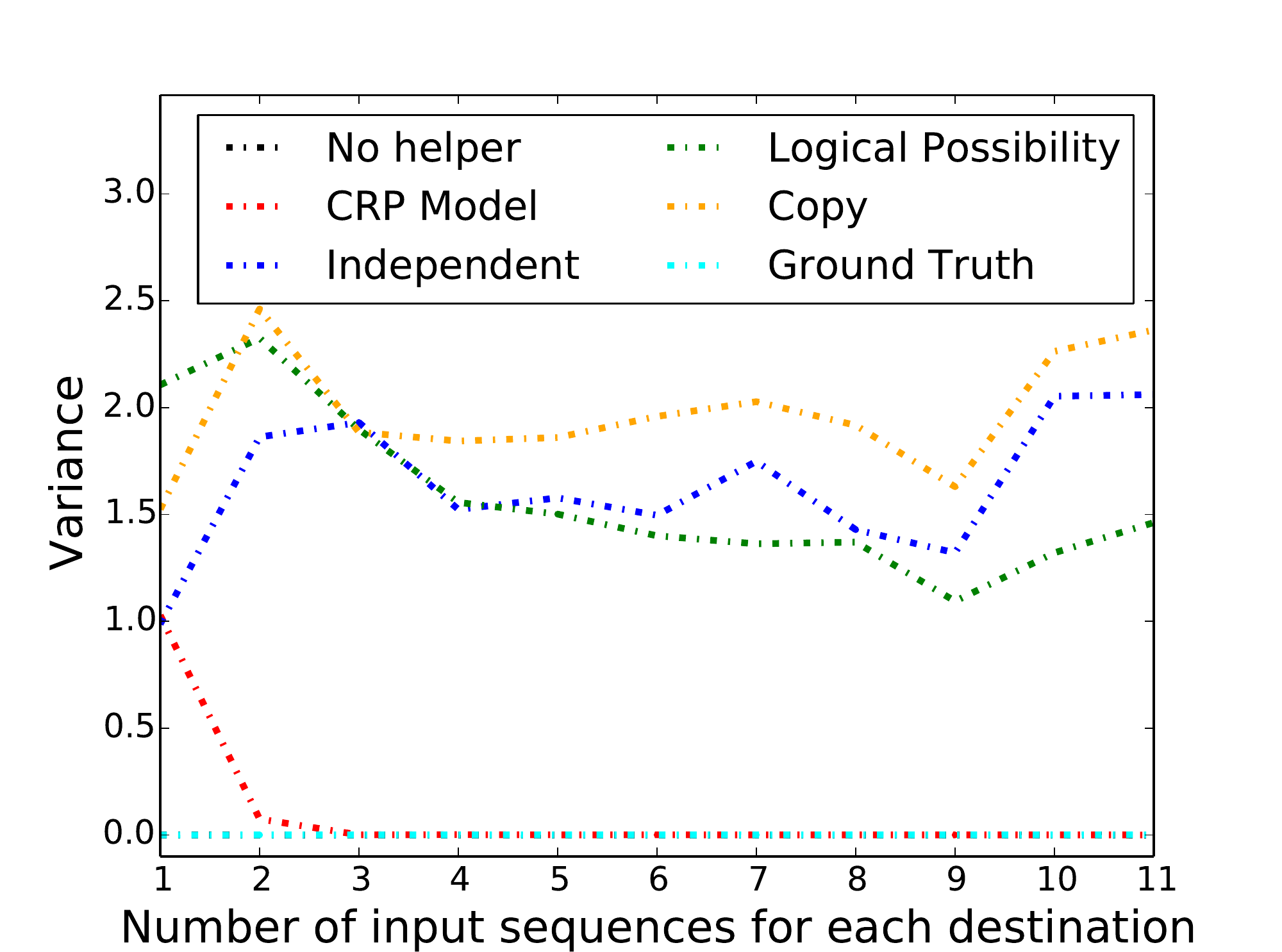}
	\caption{Model scores (a) and variance (b) as a function of the number of input sequences. Variance of No helper and Ground Truth is always 0.}
    \label{fig:exp3}
\end{figure}

We executed the test, varying the number of input sequences to evaluate the dependence of each model on the amount of input data.
Fig.~\ref{fig:exp3}(a) shows the average score of each model as a function of the number of input sequences. 
Generally, the Helper using the nonparametric Bayesian model is more beneficial than any alternative model. This model achieves good performance (almost as high as the Ground Truth model) using even a small number of input sequences.

Further, because the alternative models depend highly on the input sequences, the scores of these models are unstable. Fig.~\ref{fig:exp3}(b) shows the average variance of the results for each subgoal setting (we repeated the experiment 5 times for each subgoal setting; this score is the variance of these). None of the alternative models can provide stable user support. In contrast, the nonparametric Bayesian model provides stable support using even a small number of input sequences (over two input sequences). This stability is a key factor for user support, since unstable help can confuse and frustrate users. 

\section{Conclusion}

We presented a model of how humans infer subgoals from observations of complex action sequences. This model used rational hierarchical planning over subgoal structures generated by a nonparametric Bayesian model to capture people's intuitions about the structure of intentional actions. We showed how Bayesian inference over this generative model using a novel MCMC method predicts quantitative human subgoal inferences with high accuracy and precision. We then showed that our model is useful in practice, enhancing performance in an application to an artificial user support task.

Our modeling and experimental scenarios are extremely simplistic in comparison with real human behavior. 
One limitation is our assumption that subgoal sequences are chosen according to probabilities determined by the Dirichlet Process. More generally, the probability of choosing a particular subgoal sequence will depend on the efficiency of that subgoal sequence, relative to the alternatives. To  enhance the expressiveness of our model, Infinite PCFGs~\cite{Liang2007}, Adaptor Grammars~\cite{Johnson2007}, or fragment grammars~\cite{ODonnell2015} are promising extensions to the simple Dirichlet Process we employ. These and other frameworks which can be applied to structured goal representation complement our work here by naturally interfacing with models of hierarchical planning analogously to the model we describe.

\section{Acknowledgments}
This work was supported by the Center for Brains, Minds \& Machines (CBMM), funded by NSF STC award CCF-1231216, and by NSF grant IIS-1227495.

%\newpage

\bibliographystyle{aaai}
\small{
	\bibliography{paper,book}
}

\end{document}